\documentclass{article}

\usepackage[nonatbib, preprint]{neurips_2019}
\newcommand{\R}{{\rm I\!R}}

\usepackage[utf8]{inputenc} 
\usepackage[T1]{fontenc}    
\usepackage{hyperref}       
\usepackage{url}            
\usepackage{booktabs}       
\usepackage{amsfonts}       
\usepackage{nicefrac}       
\usepackage{microtype}      
\usepackage{graphicx}
\usepackage{amsmath}
\usepackage{algorithm}
\usepackage{algorithmic}

\title{Graph Neural Processes \\ Towards Bayesian Graph Neural Networks}

\author{%
Andrew Carr \And David Wingate
}

\begin{document}

\maketitle
\begin{abstract}
  We introduce Graph Neural Processes (GNP), inspired by the recent work in conditional and latent neural processes.
  A Graph Neural Process is defined as a Conditional Neural Process that operates on arbitrary graph data. It takes features of sparsely observed context points as input, and outputs a distribution over target points. We demonstrate graph neural processes in edge imputation and discuss benefits and draw backs of the method for other application areas. One major benefit of GNPs is the ability to quantify uncertainty in deep learning on graph structures. An additional benefit of this method is the ability to extend graph neural networks to inputs of dynamic sized graphs. 
\end{abstract}

\section{Introduction}

In this work, we consider the problem of imputing the value of an edge on a graph. This is a valuable problem when an edge is known to exist, but due to a noisy signal, or poor data acquisition process, the value is unknown. We solve this problem using a proposed method called Graph Neural Processes.

In recent years, deep learning techniques have been applied, with much success, to a variety of problems. However, many of these neural architectures (e.g., CNNs) rely on the underlying assumption that our data is Euclidean. However, since graphs do not lie on regular lattices, many of the concepts and underlying operations typically used in deep learning need to be extended to non-Euclidean valued data. Deep learning on graph-structured data has received much attention over the past decade \cite{gori-monfardini:graph-one, scarselli-yong:graph-two, li-tarlow:graph-three, zhou:graph-survey, battaglia:big-graph} and has show significant promise in many applied fields \cite{battaglia:big-graph}.  These ideas were recently formalized as \textit{geometric deep learning} \cite{bronstein2017geometric} which categorizes and expounds on a variety of methods and techniques. These techniques are mathematically designed to deal with graph-structured data and extend deep learning into new research areas.

Similarly, but somewhat orthogonally, progress has been made in Bayesian methods when applied to deep learning \cite{auld2007bayesian}. In Bayesian neural networks, uncertainty estimates are used at the weight level, or at the output of the network. These extensions to typical deep learning give insight into what the model is learning, and where it may encounter failure modes. In this work, we use some of this progress to impute the value distribution on an edge in graph-structured data.


In this work we propose a novel architecture and training mechanism which we call Graph Neural Processes (GNP). This architecture is based on the ideas first formulated in \cite{CNP:baby} in that is synthesizes global information to a fixed length representation that is then used for probability estimation. Our contribution is to extend those ideas to graph-structured data and show that the methods perform favorably.

Specifically, we use features typically used in Graph Neural Networks as a replacement to the convolution operation from traditional deep learning. These features, when used in conjunction with the traditional CNP architecture offer local representations of the graph around edges and assist in the learning of high level abstractions across classes of graphs. Graph Neural Processes learn a high level representation across a family of graphs, in part by utilizing these instructive features.

\section{Background}

\subsection{Graph Structured Data}

We define a graph $G = (a, V, E)$ where $a$ is some global attribute\footnote{typically a member of the real numbers, \R. Our method does not utilize this global attribute.}, $V$ is the set of nodes where $V = \{v_i\}_{i=1:N^v}$ with $v_i$ being the node's attribute, and $E$ is the set of edges where $E = \{e_k, v_k, u_k\}_{k=1:N^e}$ where $e_k$ is the attribute on the edge, and $v_k, u_k$ are the two nodes connected by the edge. In this work, we focus on undirected graphs, but the principles could be extended to many other types of graphs.

\subsection{Conditional Neural Processes}
First introduced in \cite{CNP:baby}, Conditional Neural Processes (CNPs) are loosely based on traditional Gaussian Processes. CNPs map an input $\vec{x_i} \in \R^{d_x}$ to an output $\vec{y_i} \in \R^{d_y}$. It does this by defining a collection of conditional distributions that can be realized by conditioning on an arbitrary number of context points $X_C : = (\vec{x_i})_{i \in C}$ and their associated outputs $Y_C$. Then an arbitrary number of target points $X_T : = (\vec{x_i})_{i \in T}$, with their outputs $Y_T$ can be modeled using the conditional distribution. This modeling is invariant to ordering of context points, and ordering of targets. This invariance lends itself well to graph structured data and allows arbitrary sampling of edges for learning and imputation.
It is important to note that, while the model is defined for arbitrary $C$ and $T$, it is common practice (which we follow) to use $ C \subset T$. 

The context point data is aggregated using a commutative operation \footnote{Works has be done \cite{kim2018attentive} to explore other aggregation operations such as attention.} $\oplus$ that takes elements in some $\R^d$ and maps them into a single element in the same space. In the literature, this is referred to as the $r_C$ context vector. We see that $r_C$ summarizes the information present in the observed context points.
Formally, the CNP is learning the following conditional distribution.
\begin{align}
P(Y_T | X_T, X_C, Y_C) \iff P(Y_T | X_T, r_C)
\end{align}
In practice, this is done by first passing the context points through a DNN $h_{\theta}$ to obtain a fixed length embedding $r_i$ of each individual context point. These context point representation vectors are aggregated with $\oplus$ to form $r_C$. The target points $X_T$ are then decoded, conditional to $r_C$, to obtain the desired output distribution $z_i$ the models the target outputs $y_i$.

More formally, this process can be defined as follows.
\begin{align}
    r_i = h_{\theta}(\vec{x_i}) && \forall \vec{x_i} \in X_C\\
    r_C = r_1 \oplus r_2 \oplus r_3 \oplus \cdots \oplus r_n \\
    z_i = g_{\phi}(\vec{y_i} | r_C) && \forall \vec{y_i} \in X_T
\end{align}

\begin{figure}
\centering
\includegraphics[scale=0.3]{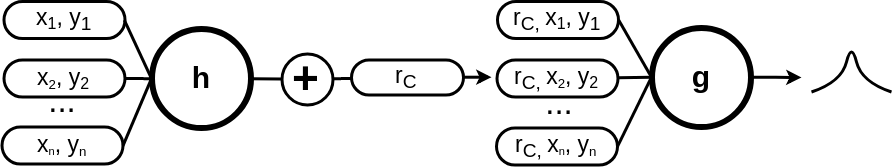}
\caption{Conditional Neural Process Architecture. The data likelihood is maximized under the output distribution. Note that $\oplus$ is an arbitrary commutative aggregation function and $h$, $g$ are parameterized neural networks. As an example, the context/target points could be $(x,y)$ coordinates and pixel intensities if an image is being modeled.}
\label{fig:cnp_plot}
\end{figure}

Traditionally, maximum likelihood is used in cases where the output distribution is continuous. In the examples explored in this work, we are dealing with categorical data and so use alternative training schemes, such as cross entropy, that are better suited to handle such outputs.

\subsection{Graph Neural Networks}

Of all the inductive biases introduced by standard deep learning architectures, the most common is that we are working with Euclidean data. This fact is exploited by the use of convolutional neural networks (CNN), where spatially local features are learned for downstream tasks. If, however, our data is Non-Euclidean (e.g., graphs, manifolds) then many of the operations used in standard deep learning (e.g., convolutions) no longer produce the desired results. This is due to the fact that the intrinsic measures and mathematical structure on these surfaces violates assumptions made by traditional deep learning operations.
There has been much work done recently in Graph Neural Networks (GNN) that operate on graph-structured inputs \cite{zhou:graph-survey}. There are two main building blocks of geometric methods in GNNs. \textit{spectral}  \cite{bronstein2017geometric} and \textit{spatial} \cite{duvenaud2015convolutional} methods. These methods are unified into \textit{Message Passing Neural Networks}, and \textit{Non-local Neural Networks}. A more thorough discussion of these topics can be found in \cite{zhou:graph-survey}. Typically, when using these GNN methods, the input dataset graphs need to have the same number of nodes/edges. This is because of the technicalities involved in defining a spectral convolution. However, in the case of Graph Neural Processes, local graph features are used which allows one to learn conditional distributions over arbitrarily sized graph-structured data.  
One important concept utilized in the spectral GNN methods is that of the Graph Laplacian. Typically, the graph Laplacian is defined as the adjacency matrix subtracted from the degree matrix $L = D - A$. This formulation, common in multi-agent systems, hampers the flow of information because it it potentially non-symmetric, and unnormalized. As such, the Normalized Symmetric Graph Laplacian is given as follows.
\begin{align}
    L =& I_n - D^{-1/2}AD^{-1/2} \label{laplacian}
\end{align}
This object can be thought of as the difference between the average value of a function around a point and the actual value of the function at the point. This, therefore, encodes local structural information about the graph itself.

In spectral GNN methods, the eigenvalues and eigenvectors of the graph Laplacian are used to define convolution operations on graphs. In this work, however, we use the local structural information encoded in the Laplacian as an input feature for the context points we wish to encode.

\section{Related Work}

\subsection{Edge Imputation}

In many applications, the existence of an edge is known, but the value of the edge is unknown (e.g, traffic prediction, social networks). Traditional edge imputation involves generating a point estimate for the value on an edge. This can be done through mean filling, regression, or classification techniques \cite{husiman}. These traditional methods, especially mean filling, can fail to maintain the variance and other significant properties of the edge values. In \cite{husiman} they show "Bias in variances and covariances can be greatly reduced by using a conditional distribution and replacing missing values with draws from this distribution." This fact, coupled with the neural nature of the conditional estimation, gives support to the hypothesis that Graph Neural Processes preserve important properties of edge values, and are effective in the imputation process.

\subsection{Bayesian Deep Learning}

In Bayesian neural networks, it is often the case where the goal is to learn a function $y = f(x)$ given some training inputs $X = \{x_1, \cdots, x_n\}$ and corresponding outputs $Y = \{y_1, \cdots, y_n\}$. This function is often approximated using a fixed neural network that provides a likely correlational explanation for the relationship between x and y. There has been good work done in this area \cite{zhang2018bayesian} and there are two main types of Bayesian Deep Learning. This is not an exhaustive list of all the methods, but a broad overview of two. 
Firstly, instead of using point estimates for the weights $W$ of the neural network hidden layers, a distribution over values is used. In other words, the weights are modeled as a random variable with an imposed prior distribution. This encodes, \textit{a priori} uncertainty information about the neural transformation. Similarly, since the weight values $W$ are not deterministic, the output of the neural network can also be modeled as a random variable. A generative model is learned based on the structure of the neural network and loss function being used. Predictions from these networks can be obtained by integrating with respect to the posterior distribution of $W$
 \begin{align}
    p(y|x, X, Y) = \int p(y|x, W)p(W|X, Y)dW.
\end{align}
This integral is often intractable in practice, and number of techniques have been proposed in the literature to overcome \cite{zhang2018bayesian} this problem. 
Intuitively, a Bayesian neural network is encoding information about the distribution over output values given certain inputs. This is a very valuable property of Bayesian deep learning that GNPs help capture.
In the case of Graph Neural Processes, we model the output of the process as a random variable and learn a conditional distribution over that variable. This fits into the second class of Bayesian neural networks since the weights $W$ are not modeled as random variables in this work.

\section{Model and Training}

\begin{algorithm}[]
\caption{Graph Neural Processes. All experiments use $n_{\text{epochs}} = 10$ and default Adam optimizer parameters}
\label{alg:algorithm}
\textbf{Require}: $p_0$, lower bound percentage of edges to sample as context points. $p_1$, corresponding upper bound. $m$, size of slice (neighborhood) of local structural eigenfeatures.\\
\textbf{Require}: $\theta_0$, initial encoder parameters. $\phi_0$ initial decoder parameters. \\
\begin{algorithmic}[1] 
\STATE Let $X$ input graphs
\FOR{$t = 0, \cdots, n_{\text{epochs}}$ }
\FOR{$x_i$ in $X$}
\STATE Sample $p \leftarrow \text{unif}(p_0, p_1)$
\STATE Assign $n_{\text{context points}} \leftarrow p \cdot |\text{Edges}(x_i)|$
\STATE Sparsely Sample $x_i^{cp} \leftarrow x_i|_{n_{\text{context points}}}$
\STATE Compute degree and adj matrix $D$, $A$ for graph $x_i^{cp}$
\STATE Compute $L \leftarrow I_{n_{\text{context points}}} - D^{-\frac{1}{2}}AD^{-\frac{1}{2}}$
\STATE Define $F^{cp}$ as empty feature matrix for $x_i^{cp}$
\STATE Define $F$ as empty feature matrix for full graph $x_i$
\FOR{edge $k$ in $x_i$}
\STATE Extract eigenfeatures $\Lambda|_{k}$ from $L$, see eq (\ref{eigen})
\STATE Concatenate $[\Lambda|_{k} ; v_k ; u_k ; d(v_k) ; d(u_k)]$ where $v_k, u_k$ are the attribute values at the node, and $d(v_k), d(u_k)$ the degree at the node.

\IF{edge $k$ $\in x_i^{cp}$}
    \STATE Append features for context point to $F^{cp}$
\ENDIF

\STATE Append features for all edges to $F$
\ENDFOR
\STATE Encode and aggregate $r_C \leftarrow h_{\theta}(F^{cp})$
\STATE Decode $\Tilde{x_i} \leftarrow g_{\phi}(F | r_C)$
\STATE Calculate Loss $l \leftarrow \mathcal{L}(\Tilde{x_i}, x_i)$
\STATE Step Optimizer
\ENDFOR
\ENDFOR
\end{algorithmic}
\end{algorithm}

While the $h_{\theta}$ and $g_{\phi}$ encoder and decoder could be implemented arbitrarily, we use fully connected layers that operate on informative features from the graph. In GNPs, we use local spectral features derived from global spectral information of the graphs. Typical graph networks that use graph convolutions require a fixed size Laplacian to encode global spectral information; for GNPs we use an arbitrary fixed size neighborhood of the Laplacian around each edge.

To be precise, with the Laplacian $L$ as defined in equation (\ref{laplacian}) one can compute the spectra $\sigma(L)$ of L and the corresponding eigenvector matrix $\Lambda$. In $\Lambda$, each column is an eigenvector of the graph Laplacian. To get the arbitrary fixed sized neighborhood around each node/edge we define the restriction of $\Lambda$ as
\begin{align}
 \Lambda|_k = (\Lambda_{kj})_{\substack{k \in r\\ 1 \le j \le m}}   \label{eigen}
\end{align}
Where $m$ is some arbitrary fixed constant based on the input space dimension. We call this restriction the local structural eigenfeatures of an edge. These eigenfeatures are often used in conjunction with other, more standard, graph based features. For example, we found that the node values and node degrees for each node attached to an edge serve as informative structural features for GNPs. In other words, to describe each edge we use the local structural eigenfeatures and a 4-tuple $(v_i, u_j, d(v_i), d(u_j))$ where $d: V \to \R$ is a function that returns the degree of a node.

\begin{figure}
    \centering
    {\includegraphics[scale=0.3]{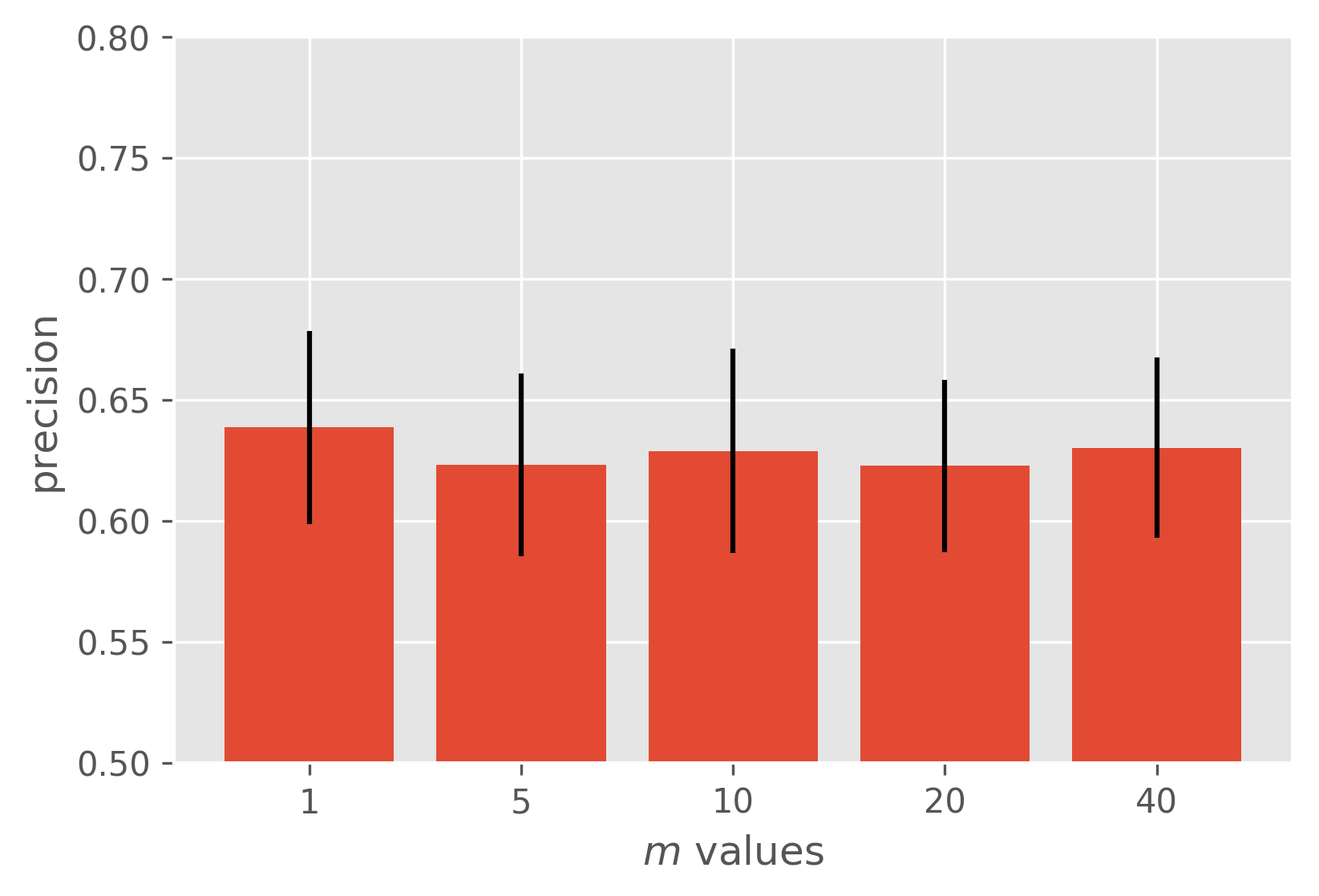}}
    {\includegraphics[scale=0.3]{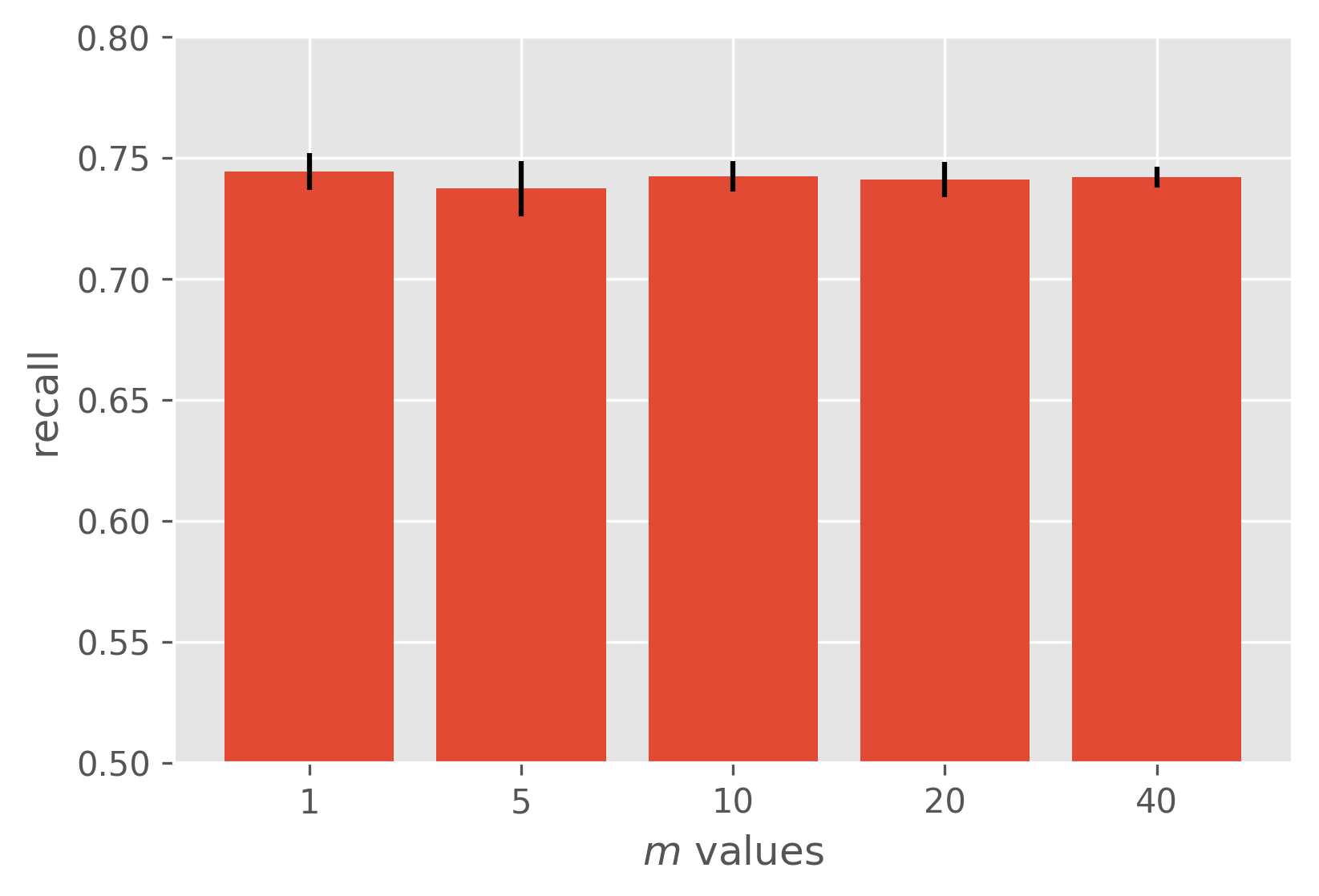} }
   {\includegraphics[scale=0.3]{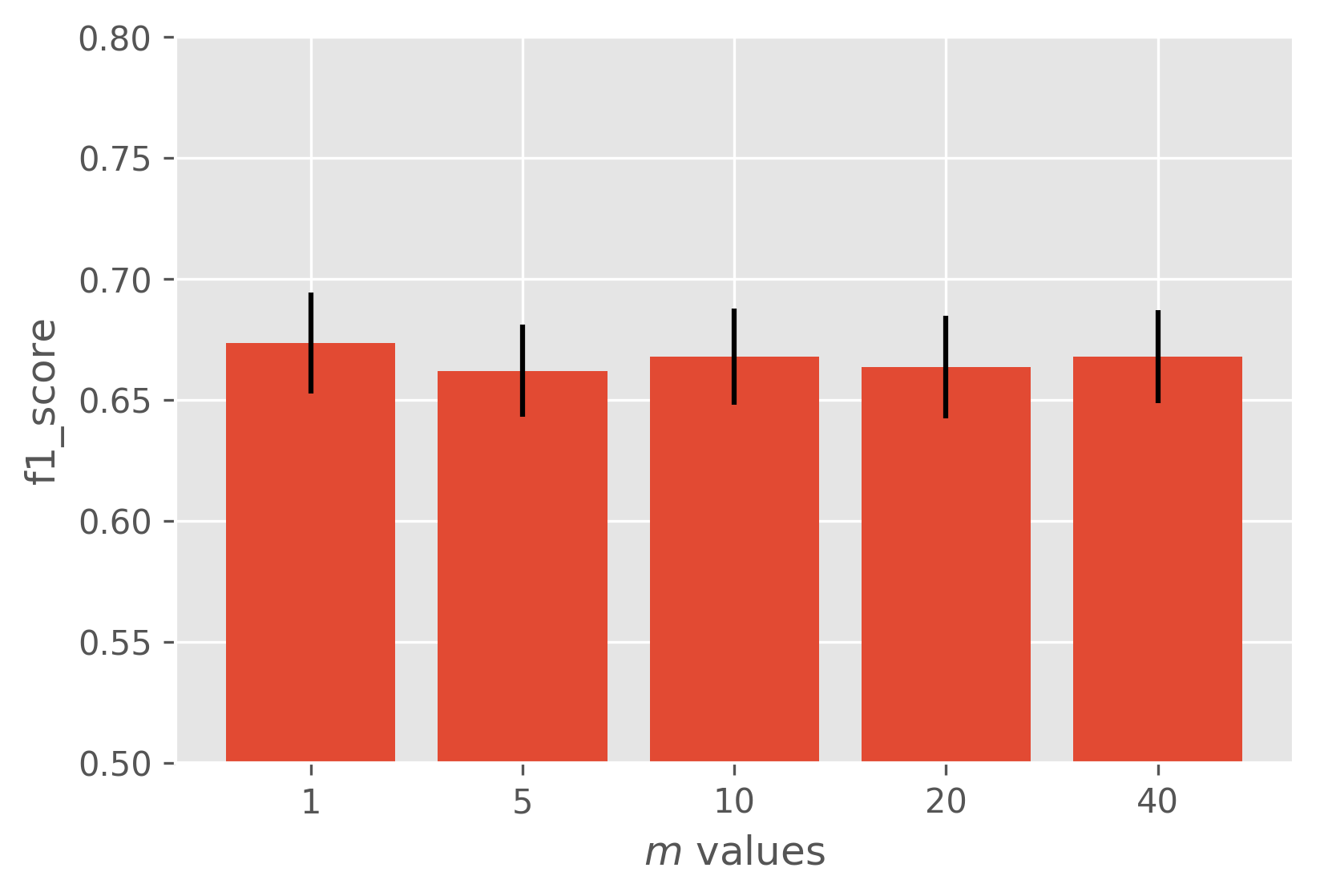} }
    \caption{Experiments with $m$ we see that the largest eigenvalue encodes sufficient information for the GNP. In other words, the performance is not statistically significant across values of $m$}
    \label{m-experiment}
\end{figure}

The introduction of these eigenfeatures also introduces an additional hyperparameter $m$ which represents the size of the neighborhood used during training. To ensure proper selection of this hyperparameter we performed a grid search over $m \in [1, n]$ where $n$ is the cardinality of the minimum set of edges of the input graphs. For this experiment we use sufficiently large graphs ($n \geq 50$) which represented approximately $\frac{1}{10}$ of the total graphs from the Tox21 AHR dataset. The results are shown in Figure (\ref{m-experiment}) and we find that the first dimension of the largest eigenvalue encodes sufficient neighborhood information for the GNP to use for learning.

Formally, we define the \textbf{Encoder} $h_{\theta}$, to be a 4-layer fully connected neural network with ReLU non-linearities. It takes the GNP features from each context point as input, and outputs a fixed length (e.g., 256) vector. This vector $r_C$ is the aggregation of information across the arbitrarily sampled context points.
The \textbf{Decoder} $g_{\phi}$ is also a 4-layer fully connected neural network with ReLU non-linearities. The decoder takes, as input, a concatenation of the $r_C$ vector from the encoder and a list of features all the edges sans attribute. Then, for each edge, the concatenation of $r_C$ and the feature vector is passed through the layers of the decoder until an output distribution is reached. Where the output distribution size is $|\cup \{e_k : e_k \in E\}|$, which is the number of unique values assigned to an edge.

Additionally, we obtain the \textbf{context points} by first defining a lower bound $p_0$ and upper bound $p_1$ for a Uniform probability distribution. Then, the number of context points is $p \cdot n$  where $p \sim \text{unif}(p_0, p_1)$ and $n = |E|$. The context points all come from a single graph, and training is done over a family of graphs. The GNP is designed to learn a representation over a family of graphs and impute edge values on a new member of that family.

The loss function $\mathcal{L}$ is problem specific. In the CNP work, as mentioned above, Maximum Likelihood is used to encourage the output distribution of values to match the true target values. Additionally, one could minimize the Kullback–Leibler, Jensen–Shannon, or the Earth-Movers divergence between output distribution and true distribution, if that distribution is known. In this work, since the values found on the edges are categorical we use the multi-class Cross Entropy.
We train our method using the ADAM optimizer \cite{kingma2014adam} with standard hyper-parameters. To give clarity, and elucidate the connection and differences between CNPs and GNPs, we present the training algorithm for Graph Neural Processes in Algorithm (\ref{alg:algorithm})

\section{Applications}

To show the efficacy of GNPs on real world examples, we conduct a series of experiments on a collection of 16 graph benchmark datasets \cite{KKMMN2016}. These datasets span a variety of application areas, and have a diversity of sizes; features of the explored datasets are summarized in Table~\ref{tab:datafeatures}.  While the benchmark collection has more than 16 datasets, we selected only those that have both node and edge labels that are fully known; these were then artificially sparsified to create graphs with known edges, but unknown edge labels.

We use the model described in section (4) and the algorithm presented in algorithm (\ref{alg:algorithm}) and train a GNP to observe $p \cdot n$ context points where $p \in [.4, .9]$ and $n = |E|$.
We compare GNPs with several baselines for the task of edge imputation. Some baselines are adapted from \cite{husiman}:
\begin{itemize}
    \item \textbf{Random}: a random edge label is imputed for each unknown edge
    \item \textbf{Common}: the most common edge label is imputed for each unknown edge
    \item \textbf{Common Neighbor}: the most common local edge label is imputed for each unknown edge, to test learning generalization over simple statistics
    \item \textbf{Random forest}: from scikit-learn; default hyperparameters
    \item \textbf{Neural Network}: network designed with comparability in mind, e.g. same number of parameters, training time, training data, optimizer as our GNP model while adhering to modern practices of hyperparameter tuning
\end{itemize}
For each algorithm on each dataset, we calculate weighted precision, weighted recall, and weighted F1-score (with weights derived from class proportions), averaging each algorithm over multiple runs.  Statistical significance was assessed with a two-tailed t-test with $\alpha=0.05$.


\subsection{Results}

\begin{figure}[ht]
\centering
\includegraphics[scale=0.30]{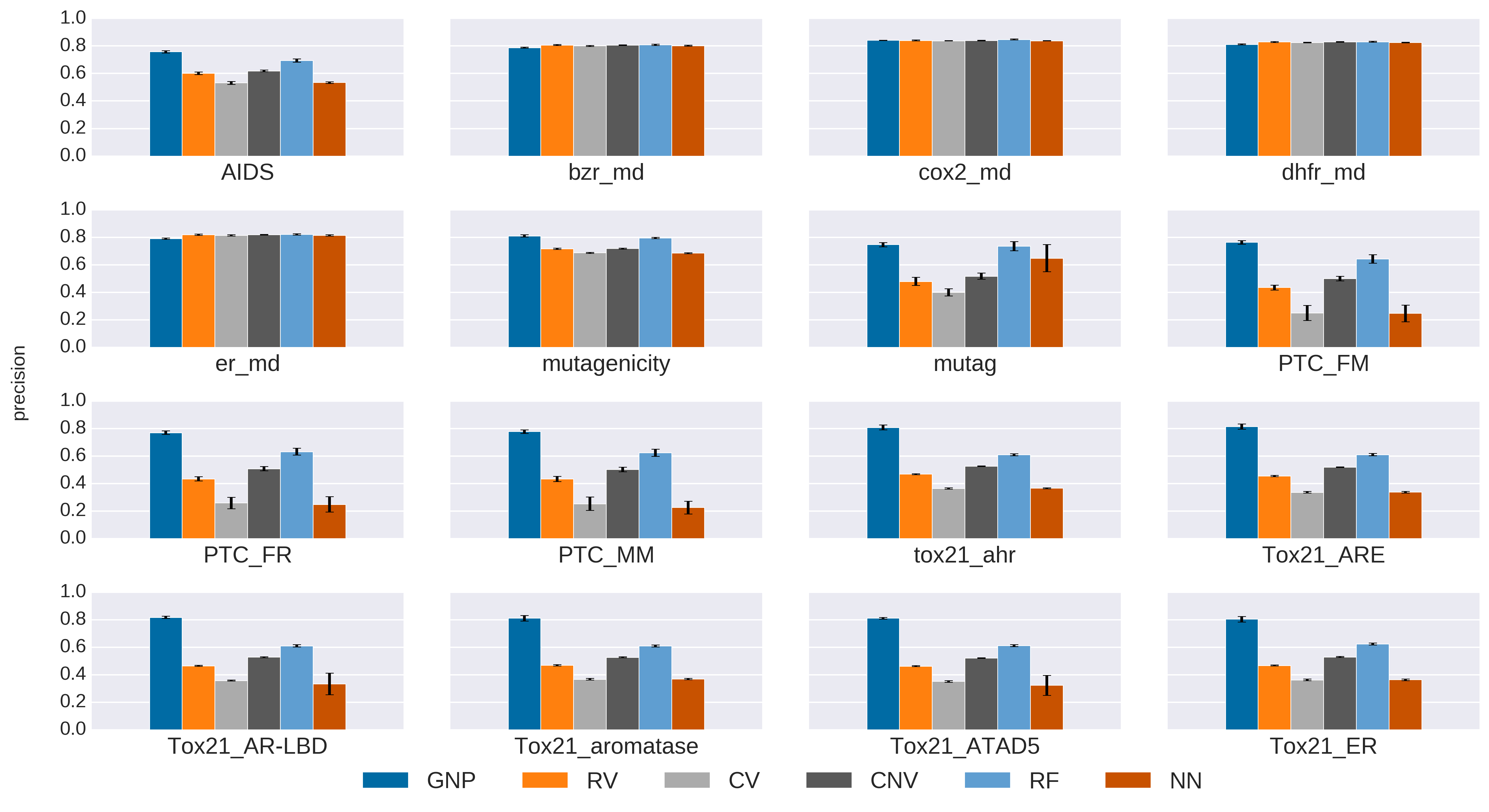}
\caption{Experimental precision graph compared with baselines, we see our method performs achieves a $\sim .2$ higher precision on average. This results holds across other metrics and across datasets which illustrates the efficacy of GNPs on the task of edge imputation.}
\label{fig:precision}
\end{figure}

\begin{figure}[ht]
\centering
\includegraphics[scale=0.30]{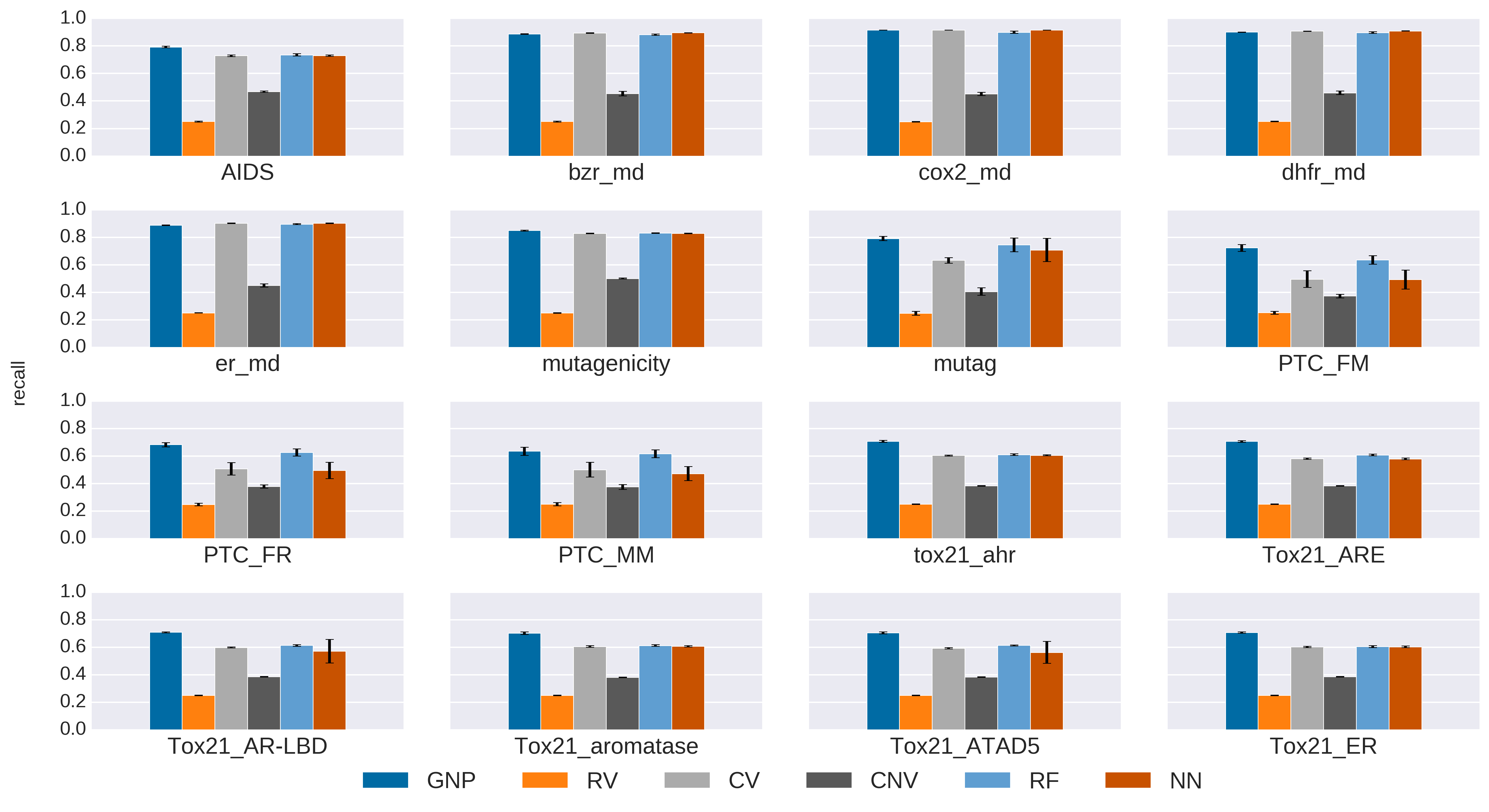}
\caption{Experimental recall graph compared with baselines}
\label{fig:recall}
\end{figure}

\begin{figure}[ht]
\centering
\includegraphics[scale=0.30]{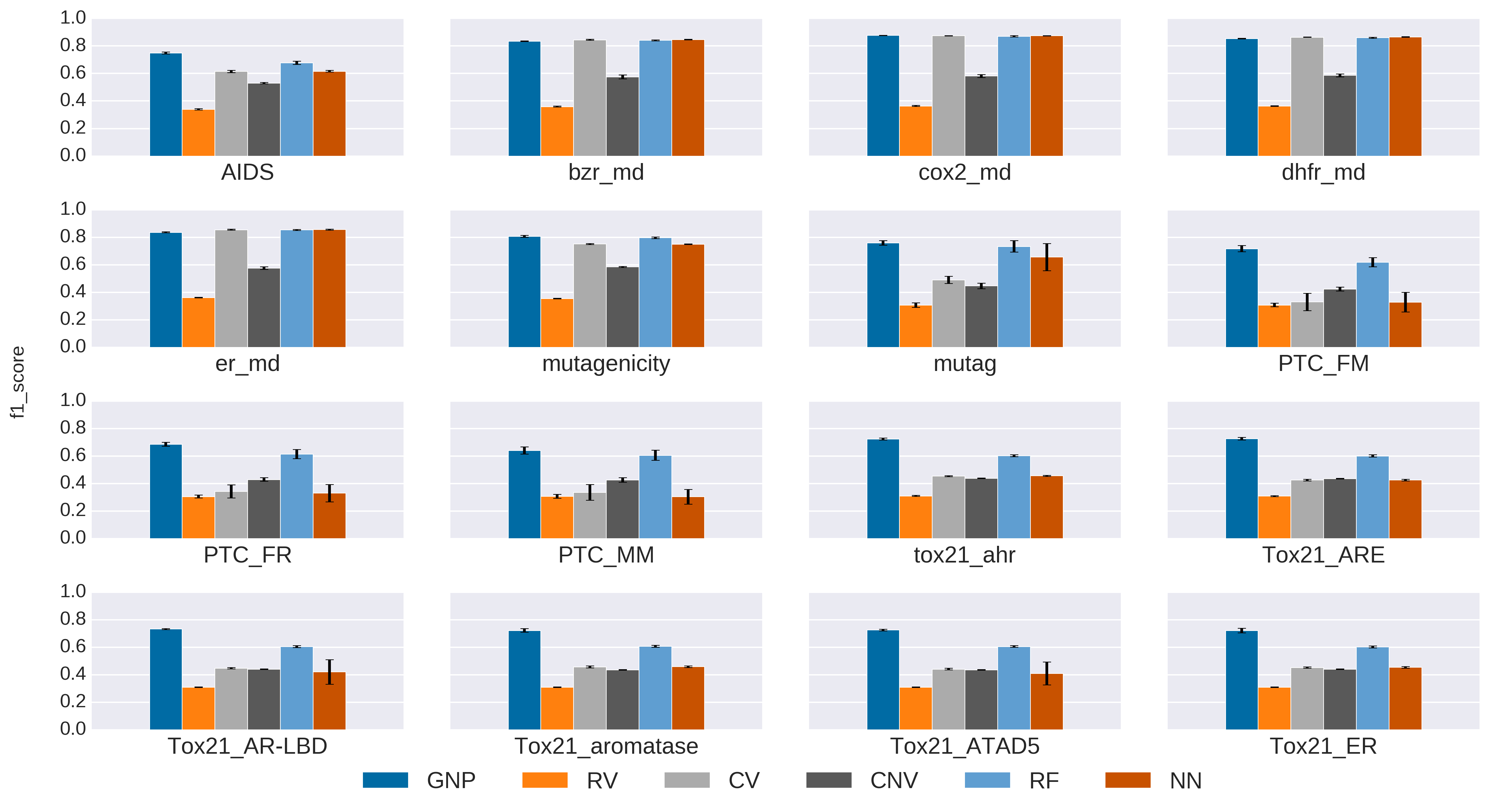}
\caption{Experimental F1-score graph compared with baselines.}
\label{fig:f1-score}
\end{figure}

The results are pictured in Figures (\ref{fig:precision} - \ref{fig:f1-score}). We first note several high-level results, and then look more deeply into the results on different subsets of the overall dataset.

First, we note that the GNP provides the best F1-score on 14 out of 16 datasets, and best recall on 14 out of datasets (in this case, recall is equivalent to classification accuracy). By learning a high-level abstract representation of the data, along with a conditional distribution over edge values, the Graph Neural Process is able to perform well at this edge imputation task, beating both naive and strong baselines. The GNP is able to do this on datasets with as few as about 300 graphs, or as many as about 9000.  We also note that the GNP is able to overcome class imbalance.

\begin{table}[ht]
\centering
\begin{tabular}{llllc}
\hline
Dataset & $|X|$ & $\bar{|N|}$ & $\bar{|E|}$ & $|\cup \{e_k\}|$ \\
\hline
AIDS       & 2000 & 15.69 & 16.20 & 3 \\
BZR\_MD    & 306 & 21.30 & 225.06 & 5 \\
COX2\_MD    & 303 & 26.28 & 335.12 & 5 \\
DHFR\_MD   & 393 & 23.87 & 283.01 & 5 \\
ER\_MD     & 446 & 21.33 & 234.85 & 5 \\
Mutagenicity   & 4337 & 30.32 & 30.77 & 3 \\
MUTAG   & 188 & 17.93 & 19.79 & 4 \\
PTC\_FM   & 349 & 14.11 & 14.48 & 4 \\
PTC\_FR   & 351 & 14.56 & 15.00 & 4 \\
PTC\_MM   & 336 & 13.97 & 14.32 & 4 \\
Tox21\_AHR   & 8169 & 18.09 & 18.50 & 4 \\
Tox21\_ARE   & 7167 & 16.28 & 16.52 & 4 \\
Tox21\_AR\-LBD	   & 8753 & 18.06 & 18.47 & 4 \\
Tox21\_aromatase   & 7226 & 17.50 & 17.79 & 4 \\
Tox21\_ATAD5	  & 9091 & 17.89 & 18.30 & 4 \\
Tox21\_ER	   & 7697 & 17.58 & 17.94 & 4 \\
\hline
\end{tabular}
\caption{Features of the explored data sets}
\label{tab:datafeatures}
\end{table}

\textbf{AIDS}.  The AIDS Antiviral Screen dataset \cite{riesen2008iam} is a dataset of screens checking tens of thousands of compounds for evidence of anti-HIV activity. The available screen results are chemical graph-structured data of these various compounds.  A moderately sized dataset, the GNP beats the RF by 7\%.

\textbf{bzr,cox2,dhfr,er}.  These are chemical compound datasets BZR, COX2, DHFR and ER which come with 3D coordinates, and were used by \cite{mahe2006pharmacophore} to study the pharmacophore kernel.
Results are mixed between algorithms on these datasets; these are the datasets that have an order of magnitude more edges (on average) than the other datasets.  
These datasets have a large class imbalance: predicting the most common edge label yields around 90\% accuracy.  For example, the bzr dataset has 61,594 in class 1, and 7,273 in the other 4 classes combined. 
Even so, the GNP yields best F1 and recall on 2/4 of these; random forest gives the best precision on 3/4.
It may simply be that there is so much data to work with, and the classes are so imbalanced, that it is hard to find much additional signal.  

\textbf{mutagenicity, MUTAG}.  The MUTAG dataset \cite{debnath1991structure} consists of 188 chemical compounds divided into two classes according to their mutagenic effect on a bacterium.
 While the mutagenicity dataset \cite{kazius2005derivation} is a collection of molecules and their interaction information with in vitro.

Here, GNP beats the random forest by several percent; both GNP and RF are vastly superior to naive baselines.

\textbf{PTC\_*}.
The various Predictive Toxicology Challenge \cite{helma2001predictive} datasets consist of several hundred organic molecules marked according to their carcinogenicity on male and female mice and rats.
\cite{helma2001predictive} On the PTC family of graphs, GNP bests random forests by 10-15\% precision, and 3-10\% F1-score; both strongly beat naive baselines.

\textbf{Tox21\_*}.  This data consists of 10,000 chemical compounds run against human nuclear receptor signaling and stress pathway, it was originally designed to look for structure-activity relationships and improve overall human health.
\cite{TOX2014}  On the Tox family of graphs, the GNP strongly outperforms all other models by about 20\% precision; about 12\% F1; and about 10\% recall.


\section{Areas for Further Exploration}

This introduction of GNPs is meant as a proof of concept to encourage further research into Bayesian graph methods. As part of this work, we list a number of problems where GNPs could be applied.
\textbf{Visual scene understanding} \cite{raposo, NIPS20177082} where a graph is formed through some non-deterministic process where a collection of the inputs may be corrupted, or inaccurate. As such, a GNP could be applied to infer edge or node values in the scene to improve downstream accuracy.
\textbf{Few-shot learning, garcia2018fewshot} where there is hidden structural information. A method like \cite{kemp2008discovery} could be used to discover the form, and a GNP could then be leveraged to impute other graph attributes.

\textbf{Learning dynamics of physical systems} \cite{battaglia2016interaction, chang2016compositional, watters2017visual, van2018relational} \cite{sanchez2018graph} with gaps in observations over time, where the GNP could infer values involved in state transitions.

\textbf{Traffic prediction} on roads or waterways \cite{li2017diffusion, cui2018high} or throughput in cities. The GNP would learn a conditional distribution over traffic between cities

\textbf{Multi-agent systems} \cite{sukhbaatar2016learning, hoshen2017vain, kipf2018neural} where you want to infer internal agent state of competing or cooperating agents. GNP inference could run in conjunction with other multi-agent methods and provide additional information from graph interactions.

\textbf{Natural language processing} in the construction of knowledge graphs \cite{bordes2013translating, onoro2017representation, hamaguchi2017knowledge} by relational infilling or reasoning about connections on a knowledge graph. Alternatively, they could be used to perform \textbf{semi-supervised text classification} \cite{kipf2016semi} by imputing relations between words and sentences.

There are a number of computer vision applications where graphs and GNPs could be extremely valuable. For example, one could improve \textbf{3D meshes and point cloud} \cite{wang2018dynamic} construction using lidar or radar during tricky weather conditions. The values in the meshes or point clouds could be imputed directly from conditional draws from the distribution learned by the GNP. 

\section{Conclusions}

In this work we have introduced Graph Neural Processes, a model that learns a conditional distribution while operating on graph structured data. This model has the capacity to generate uncertainty estimates over outputs, and encode prior knowledge about input data distributions. We have demonstrated GNP's ability on edge imputation and given potential areas for future exploration.

While we note the encoder and decoder architectures can be extended significantly by including work from modern deep learning architectural design, this work is a step towards building Bayesian neural networks on arbitrarily graph structured inputs. Additionally, it encourages the learning of abstractions about these structures. In the future, we wish to explore the use of GNPs to inform high-level reasoning and abstraction about fundamentally relational data.

\end{document}